\documentclass[11pt,a4paper]{article}
\usepackage{emnlp2021}
\usepackage{times}
\usepackage{footnote}
\usepackage{latexsym}
\usepackage{xcolor}
\usepackage{url}
\usepackage{graphicx} 
\usepackage{booktabs}
\usepackage{multirow}
\usepackage{amsmath}
\usepackage{color, soul}

\usepackage{pbox}
\usepackage{hyperref}
\usepackage{enumitem}
\usepackage{arydshln}

\usepackage{microtype}

\usepackage{todonotes}
\definecolor{deepblue}{rgb}{0.16, 0.32, 0.75}
\definecolor{indiagreen}{rgb}{0.00, 0.44, 0.00}

\title{HowSumm: A Multi-Document Summarization Dataset Derived from WikiHow Articles}

\author{
    Odellia Boni \thanks{Corresponding author} ,
    Guy Feigenblat \footnotemark ,
    Guy Lev,
    \\  \textbf{Michal Shmueli-Scheuer,
    Benjamin Sznajder, David Konopnicki \footnotemark[\value{footnote}]}
    \\
    \{odelliab, guylev, shmueli, benjams\}@il.ibm.com
    \\ IBM Research - AI }

\date{October 2021}

\begin{document}

\maketitle\renewcommand{\thefootnote}{\fnsymbol{footnote}}\footnote[2]{Work was done while author in IBM}
\renewcommand{\thefootnote}{\arabic{footnote}}
\begin{abstract}

We present \textsc{HowSumm}, a novel large-scale dataset for the task of query-focused
multi-document summarization (qMDS), which targets the use-case of generating actionable instructions from a set of sources. This use-case is different from the use-cases covered in existing multi-document summarization (MDS) datasets and is applicable to educational and industrial scenarios. 
We employed automatic methods, and leveraged statistics from existing human-crafted qMDS datasets, to create \textsc{HowSumm} from wikiHow website articles and the sources they cite. We describe the creation of the dataset and discuss the unique features that distinguish it from other summarization corpora. Automatic and human evaluations of both extractive and abstractive summarization models on the dataset reveal that there is room for improvement. 
We propose that \textsc{HowSumm} can be leveraged to advance summarization research.

\end{abstract}

\section{Introduction}
\label{intro}

Summarization has become a popular NLP task both in extractive and abstractive settings~\cite{nenkova2011}.
In recent years, progress has been very fast mainly due to the use of deep learning models starting with~\newcite{nallapati-etal-2016-abstractive}. One major obstacle for improvement in summarization tasks, especially if supervised methods are considered, is the lack of large, high-quality datasets for training summarization systems. Since summaries are difficult and expensive to produce by experts, human crafted summarization datasets contain only a few dozen instances.   

Therefore, the main technique for obtaining large summarization datasets has been to exploit texts written by humans as part of some editorial process that can be considered as summaries even if not created explicitly as such. For single document summarization, the CNN/Dailymail dataset~\cite{nallapati-etal-2016-abstractive} takes 
 advantage of key points associated with news articles as part of the article publication process. Similarly, 
Gigaword considers article titles as summaries of the corresponding article's first sentences~\cite{rush-etal-2015-neural}. 

\begin{figure}[t]
\begin{center}
  \includegraphics[width=1\columnwidth]{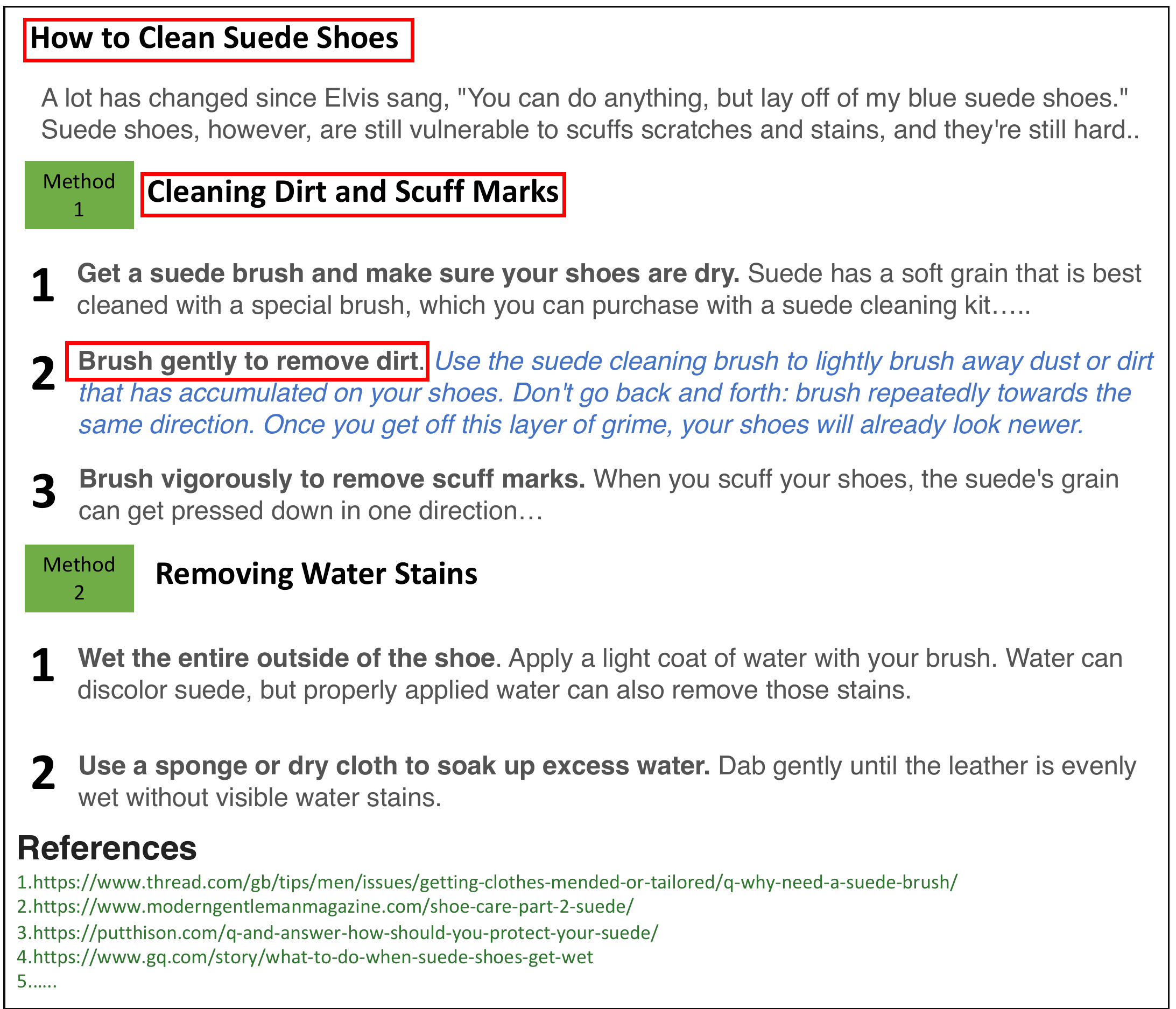}
 \end{center}
  \caption{``How to Clean Suede Shoes'' article with 2 methods. Method 1 with 3 steps, method 2 with 2 steps. 
  }~\label{fig:example}
\end{figure}

Clearly, multi-document summarization datasets (MDS), in which one's task is to summarize several source documents, are even harder to obtain. 

To tackle this challenge, several recent works have created large-scale datasets for MDS by considering existing texts to be summaries of several other sources, as described in section ~\ref{related}. 
Inspired by these works, we created a dataset based on how-to articles. 
How-to articles aim at providing information about {\it how to do something}, and can cover a variety of topics (e.g. education, medicine, or cleaning). 
Such articles range from a simple, ordered list of instructions to reach a goal, to less linear texts, outlining different ways to realize something, with arguments, conditions, etc.

wikiHow.com is a very popular website\footnote{According to the site itself, Comscore ranks wikiHow in the top 150 most visited publishers in the world.} consisting of how-to articles written by a community of authors on a wide variety of topics. The articles follow a well-defined structure that presents readers with instructions organized into {\it methods} 
and {\it steps}. 

An article contains several methods, each comprised of a title and several steps, where a step contains a title and 
a text. 
In addition, each article cites a set of sources (i.e., references), which are used by the article authors to synthesize the wikiHow article.
Figure~\ref{fig:example} shows a ``How to Clean Suede Shoes''\footnote{\url{https://www.wikihow.com/Clean-Suede-Shoes}} article, with methods, and steps. Method 1 is titled ``Cleaning Dirt and Scuff Marks'' in which the first step is titled ``Get a suede brush and make sure your shoes are dry''. 

We leverage wikiHow to create a query-focused MDS (qMDS) dataset, named \textsc{HowSumm}, by defining the references of a wikiHow article as the set of document sources to summarize, the titles as queries, and corresponding parts of the article content as target summaries. Specifically, we extract two types of target summaries from each article -- summaries originating from steps (\textsc{HowSumm-Step}), which are typically short, and those originating from methods (\textsc{HowSumm-Method}), which are longer. Consider the example in Figure~\ref{fig:example}. The text of step 2 of method 1 (in blue) is considered as the target-summary of the source documents when using as the query the title of the step and potentially also the titles of method 1 and the wikiHow article (the three titles in red boxes). Note that \textsc{HowSumm-Step} includes five such target summaries from this specific article, corresponding to the 5 (3+2) steps. Similarly, we define the concatenation of the text of all 3 steps of method 1 as a target-summary of the source documents when using as the query the title of the method and potentially also the titles of all steps under method 1, and the wikiHow article title. \textsc{HowSumm-Method} includes two such target summaries from this specific article, corresponding to the two methods.

The resulting dataset is suitable for a use-case that is significantly different than the ones considered in other summarization datasets ; 
distilling how-to instructions from the sources differs from extracting the most salient relevant information as done in News summarization, generating wikipedia paragraphs or scientific abstracts. Such a use-case, we believe, is applicable in educational and industrial applications. For example, in the technical support domain, a major task is composing troubleshooting documents that aim at solving some problem. This requires extracting the relevant problem resolution alternatives from potential sources (including, webpages, knowledge bases, white papers, etc.), and creating a coherent, actionable, and easy to follow (e.g., by using lists) instructions.

It should be noted that two large-scale summarization datasets based on wikiHow articles already exist~\cite{DBLP:journals/corr/abs-1810-09305}. One uses step title as the summary of the step's text, the other treats a concatenation of all titles in the article as summary of all other text in the article. 
These datasets target the task of generic summarization using only article's text, while we target query-focused summarization using the article sources. Thus, \textsc{HowSumm} dataset is the first qMDS dataset based on wikiHow.

Our contributions are as follows: (1) We propose a new, large qMDS dataset, containing 11,121 long summaries, and 84,348 short summaries (available at \url{https://ibm.biz/BdfhzH}) based on the wikiHow website. This dataset focuses on new use-case of generating instructions from sources. 
(2) We analyze the particularities of \textsc{HowSumm}'s target summaries and sources as opposed to some other MDS datasets. (3) We evaluate abstractive and extractive baselines on this new dataset.
 (4) Additionally, we propose a new methodology to ensure dataset quality leveraging previously existing human-crafted datasets. 
\section{Large MDS Datasets}
\label{related}
In recent years, several large-scale datasets for the task of MDS emerged 
. Multi-News~\cite{fabbri-2019-multiNews} and WCEP~\cite{wcep-2020} utilize professionally edited summaries of news events. As for other domains such human summaries are hard to obtain, an approach to consider existing texts as summaries (even if they were not written with that intention in mind) was used. WikiSum~\cite{Liu-2018-wikipedia} and auto-\textit{h}MDS~\cite{zopf-2018-auto-hmds} use lead paragraphs of Wikipedia articles as summaries of their cited reference sources. 
 Multi-XScience~\cite{Lu2020MultiXScienceAL} treats Related Work sections in scientific papers as summaries of cited articles' abstracts. For query focused setting, in the \textsc{AQuaMuSe} system~\cite{kulkarni2020aquamuse}, the authors propose a general method for building qMDS datasets based on question answering datasets where the answer serves as summary of relevant documents extracted from a corpus. They establish this approach by 
extracting answers that are Wikipedia paragraphs from Google Natural Questions~\cite{47761} and documents from Common Crawl~\cite{raffel2020exploring} corpus.

In some cases, such as Multi-XScience and Multi-news, the documents to summarize contain only cited sources. In other cases, those are extended by search results to increase coverage.

An evident challenge when creating summarization datasets in such a way, is ensuring that the information in the suggested summary indeed exists in the suggested source documents. To this end, several methods were suggested: from using only top search results on a certain topic (WikiSum), to training a relevance classifier (WCEP), ending with using text from ground truth summary in the search for sources(\textsc{AQuaMuse}, auto-\textit{h}mds).    

\section{The \textsc{HowSumm} Dataset}

\subsection{wikiHow.com}
wikiHow.com is a community website that contains more than 200K how-to articles. The average wikiHow article has been edited by 23 people and reviewed by 16 people\footnote{\url{https://www.wikihow.com/wikiHow:About-wikiHow}}. Each article aims to provide a reliabe answer a reader's ``question'' (``How to do something'').
The question is expressed in the article's title. The article itself begins with an introductory paragraph followed by a list of methods. Each method has a title, and is divided into several steps. Each step begins usually with a text in bold (which we treat as the step's title), followed by a regular text (which we consider as the step's text). The article ends with a list of references, related articles and Q\&A. 
In addition, each article includes metadata such as creation date, last edition date, \# of authors, category (out of 20) and subcategories.

When writing a wikiHow article, one should follow some guidelines\footnote{\url{https://www.wikihow.com/Write-a-New-Article-on-wikiHow}}. Specifically, two important aspects are emphasized, the \textit{content} of the article, and its \textit{sources} (i.e., references). The content should includes facts, ideas, methodologies, and examples. Short background information is also acknowledged. In order to create an accurate and authoritative article, authors are encouraged to use sources. The sources should be reliable, and should not include websites like Wikipedia, web forums, blogs, advertisements, ask.com, etc. However, to avoid legal issues, authors should use their own words to express information from the sources.

\subsection{Dataset Instances Creation}

\begin{figure*}
\begin{center}
  \includegraphics[width=1\textwidth]{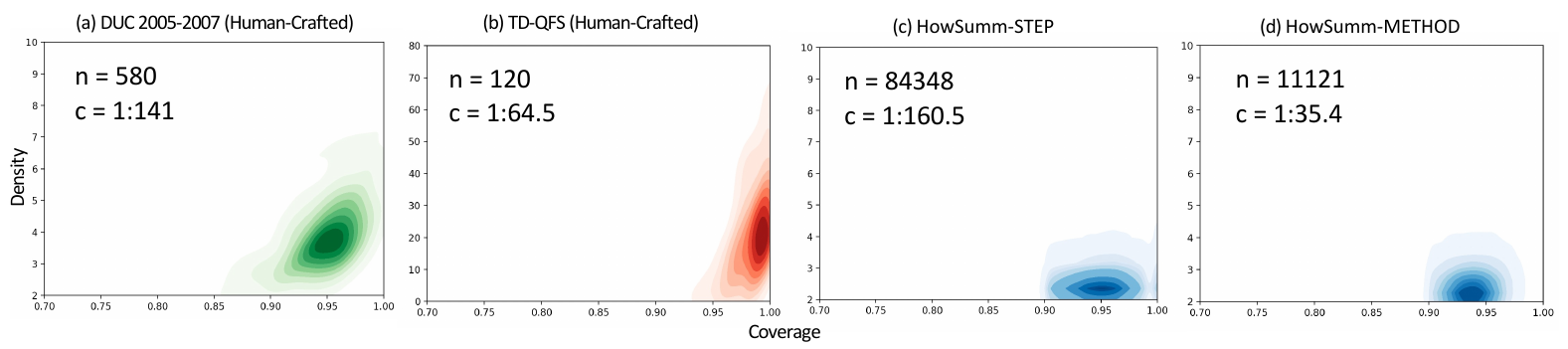}
 \end{center}
  \caption{Coverage and density in qMDS datasets. ``c'': average compression ratio; ``n'': number of instances.}~\label{fig:density-coverage}
\end{figure*}

Our goal is to use wikiHow article's content as 
query-focused summaries of the sources. 
For each  wikiHow article, we consider several target summaries, one per method (\textsc{HowSumm-method}), and one per step (\textsc{HowSumm-step}).
We exclude the option of using entire wikiHow articles as target summaries
because of the complex structure and length of target summaries in this case.

We crawled wikiHow for its articles in English, 
and, using boilerpy3\footnote{\url{https://pypi.org/project/boilerpy3/}}, extracted their structured content 
including titles, and their sources URLs. We use regular expressions to remove misleading terms from article title such as ``With Pictures'' or ``In 10  Steps''. We divide each step text into a title (bold text at the start) and a target summary (the plain text that follows). We omitted steps where a title or a target summary could not be found. To create target summaries for a method, we concatenate target summaries of its steps. 

Each instance of the dataset contains the sources URLs, the target summary and corresponding article title, method title and step titles (single for \textsc{HowSumm-Step}, multiple for \textsc{HowSumm-Method}). This setting allows to experiment with several titles combinations as query. For example, for a step, the most obvious choice for a query is the step title, but the combination of the article title, the respective method title and the step title can also serve as the query. Similarly, for methods, in addition to method title, the combination of method title and the titles of the steps comprising the method can also be used as a query.     

To overcome the problem of sources changing over time,
we provide archived sources URLs obtained by using the article's ``published date'' in Archive.org API request.

\subsection{Dataset Instances Filtering}
For some articles the content was updated and is no longer consistent with the sources or the sources are only partially accessible.
Therefore, we need further filtering to get 
high-quality instances, suitable for summarization. 

To this end, we employ metrics of \textit{coverage}, \textit{density}, and \textit{novel n-grams}. 
These metrics were first proposed in \newcite{grusky-etal-2018-newsroom} 
to characterise the level of extractiveness in generic single document summarization datasets. Coverage evaluates how much of the summary text originates from the source, while density indicates the length of shared texts between the summary and the source. Novel n-grams measures level of variance between the summary and the source's text.  \newcite{fabbri-2019-multiNews} applied the same metrics for generic MDS datasets by concatenating source documents. Since these metrics rely solely on common text fragments between a summary and the corresponding sources (and on summary length), we can adopt them also for 
query-focused summarization.

We first studied these metrics 
for existing, human created qMDS datasets, namely, \textsc{DUC 2005-2007}~\cite{10.1016/j.ipm.2007.01.019} and \textsc{TD-QFS}~\cite{baumel-2016-td-qfs}. Both datasets introduce multiple summaries per query, and we treated each of these summaries as an additional instance during metrics calculation. Thus, we ended up with 580 and 120 data points for \textsc{DUC 2005-2007} and \textsc{TD-QFS}, respectively. Figures~\ref{fig:density-coverage}(a) and ~\ref{fig:density-coverage}(b) show density and coverage distributions, and table~\ref{tab:novelty-n-grams} presents novel n-grams percentage across the human created qMDS datasets. Both datasets have similarly high coverage values, while \textsc{TD-QFS} has higher density and lower percentage of novel n-grams, implying a more extractive nature of its summaries. 

Assuming these human created datasets are representative of high quality instances, we use their metrics bounds as thresholds for filtering instances for the \textsc{HowSumm} dataset. 
Specifically, we require that each instance from our dataset will attain coverage $>$ 0.9, density $>$ 2, and novel bi-grams $<$ 0.6. This filtering process yielded 84,348 instances for \textsc{HowSumm-Step}, 11,121 instances for \textsc{HowSumm-Method}.
The average number of target summaries stemming from the same wikiHow article in \textsc{HowSumm-Step}  and \textsc{HowSumm-Method} is 3.7 and 1.7 , respectively. 

Figures~\ref{fig:density-coverage}(c) and ~\ref{fig:density-coverage}(d) and table~\ref{tab:novelty-n-grams} contain the metrics for the resulting \textsc{HowSumm} dataset.
 
\begin{table}[thbp]
\centering
\resizebox{0.95\columnwidth}{!}{
\begin{tabular}{l|c|c|c}
Dataset & uni-grams &bi-grams &tri-grams \\ \hline \hline
\textsc{DUC 2005-2007} (Human-Crafted) &  10.1 & 46.7 & 70.9 \\
\textsc{TD-QFS} (Human-Crafted) & 2.8  &13.8 &26.6    \\
\hline
\textsc{HowSumm-step} & 9.8 & 47.3 & 78.9\\
\textsc{HowSumm-method} & 15.2 & 52.7 & 81.9 \\
\hline
 \hline
\end{tabular}
}
\caption{Novel n-grams percentage in qMDS datasets.}
\label{tab:novelty-n-grams}
\end{table} 
 
 \subsection{Human Evaluation}
We performed a human evaluation study to validate the quality of the \textsc{HowSumm} dataset. 
First, one expert identified for each sentence in a target summary, a \textit{similar} sentence in the sources, where similar is defined as ``sentences conveying the same idea not necessarily in the same words''. Then, a second expert reviewed the resulting pairs. Only pairs confirmed by the second expert were considered for the coverage calculation below. 

In total, 31 \textsc{HowSumm-Step} summaries and 15 \textsc{HowSumm-Method} summaries (with a total of 58 steps)
were annotated. $88\%$ of the pairs identified by the first expert were verified by the second one. We define \textit{coverage} of a target summary as the number of sentences in the summary having similar sentences in the sources, divided by the number of summary sentences. 
For both \textsc{HowSumm-Step} and \textsc{HowSumm-Method} the average \textit{coverage} was $0.62$ , similar to that reported for \textsc{AQuaMuSe}, another automatic qMDS dataset.


\subsection{Dataset Statistics and Analysis} 

\begin{table*}[th]
\centering\
\resizebox{0.95\textwidth}{!}{
\begin{tabular}{l|c|c|ccc|ccc}
 \textbf{Dataset}  & \textbf{Size} & \textbf{\#Sources (mean)} &\multicolumn{3}{|c|}{\textbf{Source}} & \multicolumn{3}{|c|}{\textbf{Summary}} \\
 & && \#words & \#sentences & vocab size  &  \#words & \#sentences & vocab size \\ \hline \hline
 \textsc{DUC 2005-2007 (human-crafted)}* & 145  & 27.37  & -  & 26.41  & -  & 250
& -  & -\\
\textsc{TD-QFS (human-crafted)}* &40  &  190.15 &98.67   &  8.14 & -  & 117.48  & -
& -\\
\hline
\textsc{AQuaMuSe}* & 5519 &6  & 1597.1  & 66.4  & -  & 105.9  & 3.8  & -  \\
\textsc{HowSumm-Step} & 84348 & 9.98 & 1357.37 & 66.47 & 1175198 & 98.98 & 5.23 & 70153
\\
\textsc{HowSumm-Method} & 11121 &11.19& 1455.52 & 71 & 653509 & 539.11 &31.33 & 58967  \\
\hline
 \hline
\end{tabular}}
\caption{Descriptive statistics of \textsc{HowSumm} and existing qMDS datasets. *As reported in the original papers.}
\label{tab:stat}
\end{table*}

Table~\ref{tab:stat} compares \textsc{HowSumm} with other qMDS datasets. Notice that, in addition to DUC 2005-2007 and TD-QFS which are human annotated datasets, we also report on \textsc{AQuaMuSe} which like \textsc{HowSumm} was created automatically.
Clearly, \textsc{HowSumm} is the largest dataset. 

Unlike the human created datasets which contain many short sources per instance, the sources in the automatically created datasets are fewer but much longer. Next, we look deeper into these sources.

\subsubsection{Sources}\label{sec:sources}
As described in the authoring guidelines, the sources of a wikiHow article should support the article content and increase its authority and credibility. 
Since the sources describe procedures, methodologies and explanations, they are likely to be rich in \textit{headers} to organize the concepts, \textit{Q\&A} content that provides authoritative information, in addition to \textit{lists}, which are often used to record instructions\footnote{\url{https://pressbooks.bccampus.ca/technicalwriting/chapter/lists/}}. To confirm this assumption, we randomly selected 10K sources from our dataset, and, for each one, extracted from its HTML format both Q\&As and lists (as described in~\cite{orgFAQ}). The average number of Q\&As for a source, and the average number of lists, are 2.3, and 10.3, respectively. 

We also counted HTML headers per source, and number of different headers levels \footnote{HTML supports 6 levels of headers.} used in sources that contain headers. The average number of headers for a source, and the average number of header levels, are 3.1, and 3.3, respectively. 

As a comparison, we applied the same analysis to sources from other summarization datasets, Newsroom~\cite{grusky-etal-2018-newsroom} and auto-\textit{h}MDS~\cite{zopf-2018-auto-hmds} where the source documents are news articles and Wikipedia references, respectively. Both datasets provide URLs of their sources, so sources content and their HTML format  can be reconstructed.
News sources have an average of 0.5 Q\&As , 0.2 lists, 0.3 headers per page, and use on average 1.1 different header levels. Wikipedia cited references contain an average of 0.9 Q\&As , 1.5 lists, 7.4 headers per page, and 1.6 different header levels. This confirms that \textsc{HowSumm} sources are significantly richer in content organization aids.

\subsubsection{Summaries}\label{sec:summaries}

Summaries in our dataset are how-to guides. Hence, we expect them to include {\it actionable} statements. 
In addition, following authoring guidelines, we expect summaries to include many \textit{examples}, which demonstrate readers how to apply a certain guideline. To capture the actionable aspects of the summaries, we used a rule-based approach based on POS analysis\footnote{\url{https://github.com/vaibhavsanjaylalka/Action-Item-Detection}}. In total, 35.4\% of the sentences in our dataset summaries are actionable sentences. To capture the examples, we matched a list of phrases which are synonyms of ``for example''. Here, 10.3\% of the summaries include at least one example.

We run the same analysis on two other summarization datasets having summary lengths similar to ours. 
News summaries from
Multi-News~\cite{fabbri-2019-multiNews} include only a small fraction of actionable sentences (5.12\%), and 4.19\% contain examples. Wikipedia paragraphs which serve as target summaries in auto-\textit{h}MDS~\cite{zopf-2018-auto-hmds} contain 18\% of actionable sentences, and only 1.7\% of them include examples.
\section{Experiments}\label{exp}
\subsection{Setup} 
All wikiHow articles producing \textsc{HowSumm} were split into training, validation and testing sets (80\%, 10\%, 10\%, respectively) inducing a similar split to \textsc{HowSumm-Step} and \textsc{HowSumm-Method}. In the following section, we describe the performance of different summarization models on the testing sets. Based on the average length of steps and methods texts,
we defined length limits to be 90 and 500 words, respectively.
As evaluation measures we use ROUGE scores \cite{lin-2004-rouge}\footnote{with flags -f A -a -m -n 4 -t 0 -2  4 -c 95 -u -r 1000 -p 0.5}. We follow the recommendation in \cite{wcep-2020} to evaluate models with a dynamic output length 
on full-length outputs 
, while forcing controlled length models (e.g. extractive) to return untruncated sentences up to the given length limit.

\subsection{Models}
We utilized different extractive baselines to serve as lower and upper bounds to models performance, as well as some state-of-the-art (SOTA) models.
\newline
\textbf{Lower bound.} 
 (i) \textsc{Random}: random sources sentences are chosen for the output until length limit is reached. (ii) \textsc{GreedyRel}: the source sentences with the highest Jaccard similarity to the query's bag-of-words representation comprise the output.  
\newline
\textbf{Upper bound.} 
We use two oracles. In both, each target summary sentence is matched with the most similar source sentence (from unmatched sentences pool). The resulting set of matched source sentences serves as the output. Note that the oracles do not impose a length limit. 
\textsc{Oracle-BOW} where we use Jaccard smilarity between sentences' bag-of-words representations, and \textsc{Oracle-BERT} where we apply cosine similarity between sentences' BERT embeddings\footnote{https://huggingface.co/sentence-transformers/bert-base-nli-mean-tokens}. 
\newline
\textbf{SOTA-extractive.} (i) \textsc{LexRank}: an algorithm that computes graph-based centrality score of sources sentences \cite{lexrank}. 
A query-bias of $0.7$ adjusts the algorithm to query-focused summarization \cite{otterbacher-etal-2005-using}. 
(ii) \textsc{CES}: an unsupervised model optimizing a set of features of the selected output using cross-entropy method \cite{CES}.  
\newline
\textbf{SOTA-abstractive.} We use models derived from \textsc{HierSumm}~\cite{liu-lapata-2019-hierarchical}, a neural encoder-decoder model which aims to capture inter-paragraph relations by leveraging a hierarchical Transformer \cite{vaswani2017attention} architecture.
The input for this model are the query and several dozens top ranked paragraphs taken from the source documents. The authors in \cite{liu-lapata-2019-hierarchical} split source documents into paragraphs using line-breaks, and train a neural regression model to rank them, using their ROUGE-2 recall against the target summary as ground-truth scores. As this paragraph ranker is not publicly available, we used two alternatives: a ranker which uses a BM25 algorithm \cite{10.1561/1500000019} to rank the source paragraphs given a query; and an oracle ranker which ranks the paragraphs according to their ROUGE-2 recall against the target summary. The corresponding summarization models are \textsc{BM25-HierSumm} and \textsc{Oracle-HierSumm}, respectively.

In each experiment, the \textsc{HierSumm} model was initialized with the parameters of the publicly-available pretrained model of \cite{liu-lapata-2019-hierarchical}.
We used the default training hyperparameters and fine-tuned our models for 150,000 steps.
We utilize those abstractive models only for \textsc{HowSumm-Step}, as \textsc{HowSumm-Method}  requires producing quite long outputs. 

\subsection{Query Choice}\label{sec:q_effect} 
As mentioned before, \textsc{HowSumm} allows to define several queries for the same target text. In table~\ref{tab:q_choice} we show ROUGE-1 Recall results of various models ran with different query choices for \textsc{HowSumm-Step} and \textsc{HowSumm-Method}. We see that adding titles of lower level (steps titles for \textsc{HowSumm-Method}) 
improves output score while adding title of higher level (article title for \textsc{HowSumm-Method}, article and method titles for \textsc{HowSumm-Step}) has no or negative impact. 
A possible explanation is that lower level titles act as a detailed description of required content and help focus the output, while higher level titles are more general and may cause output to drift. Therefore, 
we use step title as query for \textsc{HowSumm-Step}, and method title together with its comprising steps titles as query for \textsc{HowSumm-Method}. Appendix \textit{A} contains examples of models output.   

\begin{table*}[!h]
\centering\
\resizebox{0.85\textwidth}{!}{
\begin{tabular}{l|l|c|c|c|c}
Dataset & Query      & \textsc{GreedyRel} & \textsc{LexRank} &  \textsc{CES} & \textsc{BM25-HierSumm} \\  \hline \hline
\textsc{HowSumm-Step}   & step title &30.1 & 39.6 & 39.3&  22.3   \\
        & step + method titles &30.3 & 38.2 & 38.3& 23.0  \\
        & step + method + article titles &30.1 & 36.3 & 37.0 &  21.9  \\
        \hline\hline
\textsc{HowSumm-Method} &  method title &43.4 & 47.7 & 48.4&     -      \\
        & method + article titles &42.3 & 47.1 &48.3 &          -       \\
        &  method + article + steps titles &48.6 &53.5 & 52.2 &  -     \\
       \hline\hline
       \end{tabular}}
\caption{ROUGE-1 Recall results of \textsc{HowSumm-Step} and \textsc{HowSumm-Method} for different query choices.}
\label{tab:q_choice}
\end{table*}

\subsection{Automatic Evaluation}\label{sec:results1}
Table~\ref{tab:rouge_all} details ROUGE scores of the different models outputs. As expected, selecting random sentences yields the lowest scores. \textsc{GreedyRel}'s simple model that prioritizes similar-to-query sentences improves significantly over the random model. \textsc{LexRank} algorithm performs even better since it considers the pool of all sentences when selecting the central ones, thus avoiding redundancy. \textsc{CES} outperforms LexRank as it optimizes also additional features such as output focus. The big gap between extractive models' and extractive oracles' scores implies that there is room for improvement.

BERT-based oracle's output is characterized by higher recall but lower precision than that of BOW-based oracle. The reason for that is that while the two oracles' outputs have the same number of sentences, BERT-based oracle tends to select longer sentences (by approximately 20\%).  
This is because BERT embeddings are richer for longer sentences, while Jaccard similarity penalizes sentences pairs of different lengths.

The \textsc{HierSumm} model, with a BM25 paragraph ranker, exhibits scores similar to or lower than the extractive models. With Oracle ranker, \textsc{HierSumm} achieves scores much higher than the extractive models, but still lower than those of the extractive oracles. \textsc{HierSumm}'s results manifest the crucial role of ranking source paragraphs in this model, and show that once this ranking is adequate, \textsc{HierSumm} outperforms SOTA extractive models.
Note that \textsc{HierSumm}'s outputs are well under the length limit, with \textsc{BM25-HierSumm} creating shorter outputs than \textsc{Oracle-HierSumm} (average of 46.3 words opposed to 64.0).

\subsection{Manual Evaluation}\label{sec:results2}

In addition, we evaluated models outputs manually for randomly chosen 20 instances from \textsc{HowSumm-Step} in the following way. First, an expert provided a set of questions for each instance (based on target text). Then, for every instance and for every model, 3 annotators determined whether the questions corresponding to the instance can be answered using the model's output. Majority vote determined answerability of each question. The ratio of answerable questions is the output's information score. The annotators also rated each output on a scale of 1(poor) to 5(good) for several linguistic quality aspects using a set of questions developed in DUC \cite{10.1016/j.ipm.2007.01.019}: Is the text ungrammatical? Does it contain redundant information? Are the references to different entities clear? Is the text coherent? An average of annotators ratings determines output score for each aspect. 

Human evaluation results appear in table~\ref{tab:manual-eval}. Agreement between annotators on information questions, measured by Fleiss’s Kappa, was 0.49, indicating a moderate agreement. Table~\ref{tab:manual-eval} reveals that abstractive model outputs are more fluent compared to extractive models as higher focus, coherence and referential clarity indicate, but suffer from repetitions and low informativeness. \textsc{LexRank} and \textsc{CES} outputs achieve similar scores in most aspects. \textsc{LexRank} outputs were judged to be more informative, but with higher redundancy than those of \textsc{CES}. All models achieved low coherence score.   

\begin{table*}[!h]
\centering\
\resizebox{0.85\textwidth}{!}{
\begin{tabular}{l|l|r|r|r|r|r|r|r|r}
Dataset & Model      & \multicolumn{2}{c}{R-1} & \multicolumn{2}{c}{R-2} & \multicolumn{2}{c}{R-L} & \multicolumn{2}{c}{R-SU4}  \\
        &             & Recall & F-1      & Recall & F-1       & Recall & F-1       & Recall & F-1          \\
        \hline \hline
\textsc{HowSumm-Step}   & \textsc{Random}   &     23.0     &   22.4 &    2.7    &  2.6              &   16.8     &      16.1          &    6.5    &      6.2             \\
        & \textsc{GreedyRel}   &   30.1     &     28.7     & 7.5        & 6.8                & 21.9       & 20.5               &    11.1    &  10.3                 \\
         & \textsc{LexRank}     &    39.6    &     31.5       &   10.7    &   7.8             &     24.0     &      18.3        &  15.4      &       11.5            \\
        & \textsc{CES}        &   39.3     &     32.4    &   9.1     &      7.2          &  28.2       &     22.8          &   14.1     &      11.3             \\
        & \textsc{BM25-HierSumm}  &  22.3&     27.4   &         6.7       &    8.2    &     15.7           &   19.1     &     8.6           &   10.6                      \\
           & \textsc{Oracle-HierSumm}    &  35.6      &     40.6           &    13.9    &      15.8          &   23.8     &         26.9       &   15.8     &           18.0        \\
        & \textsc{Oracle-BOW}  &    46.0    &   47.0             &   16.9    &   17.1              &   33.1     &    33.7            &   20.4   &      20.6               \\
        & \textsc{Oracle-BERT} &   46.8     &    42.6    &    15.2    & 13.8              &   33.0     &    30.0     &   19.6     &    17.8              \\ 
          \hline\hline
\textsc{HowSumm-Method} & \textsc{Random}      &     41.5&            39.6 &8.4& 7.9& 27.8& 26.3& 15.0   &14.3                      \\
        & \textsc{GreedyRel}   &     48.6&46.3 &14.2 &13.2 & 30.5& 28.7  &20.8           & 19.6                         \\
        & \textsc{LexRank}     &   47.7     &  42.0              & 12.0       &   10.2             &  19.7      &     17.0           &    19.2    &       16.7            \\
        & \textsc{CES}        &    52.2&      46.4 &14.5 & 12.7 & 34.1& 29.9&21.9   & 19.4  \\
           & \textsc{Oracle-BOW}  &     58.9 & 59.5& 21.2& 21.2& 38.4&38.7 &27.6    & 27.7\\
        & \textsc{Oracle-BERT} &   63.2     &    56.1&  21.1    & 18.7              & 40.9     &   36.3      &  28.7    &    25.4              
        \\ \hline\hline

\end{tabular}}
\caption{ROUGE results of \textsc{HowSumm-Step} and \textsc{HowSumm-Method} for different models.}
\label{tab:rouge_all}
\end{table*}

\begin{table*}[h]
\centering\
\resizebox{0.85\textwidth}{!}{

\begin{tabular}{l|c|c|c|c|c|c}
Model      & Information & Grammaticality & Non-redundancy & Referential Clarity & Focus&Coherence   \\      
        \hline \hline

 \textsc{LexRank} &  31.08\%     &    3.93    &    3.65    & 4.15               &   3.45     &    2.65         \\
 \textsc{CES}    &  27.58\%  &    3.70     &    4.25    & 4.43                &   3.37     &    2.68          \\
 \textsc{BM25-HierSumm} &25.44\% &3.88 & 3.00 &4.53 & 4.05&3.09 \\
                                 
\end{tabular}}
\caption{Human evaluation of a sample of \textsc{HowSumm-step} for different models.}
\label{tab:manual-eval}
\end{table*}

\section{Discussion}

\subsection{Actionability Effect}
We observed in section~\ref{sec:summaries} that \textsc{HowSumm} target summaries 
contain many actionable sentences. We tried to utilize this feature; In \textsc{GreedyRel} model, we experimented with several ways to prioritize actionable sentences when creating outputs. However, all experiments resulted in a decrease of ROUGE scores.
To further investigate the issue we analyzed actionability of similar sentences in target summary and sources. To determine such pairs, we used Jaccard similarity of sentences' bag-of-words representation. Out of $2400$ pairs of such similar sentences, in $504 (21\%)$ target sentence was actionable, while matching source sentence was not. For example, source sentence ``People with any of these symptoms should see their physician'' became ``See an eye doctor for certain symptoms.'' in target text.
This indicates that humans do not consider actionability of sentence during content selection, but rather transform some selected sentences into actionable ones during summary construction.

\subsection{Examples effect}
\begin{figure}[t]
\begin{center}
  \includegraphics[width=1\columnwidth]{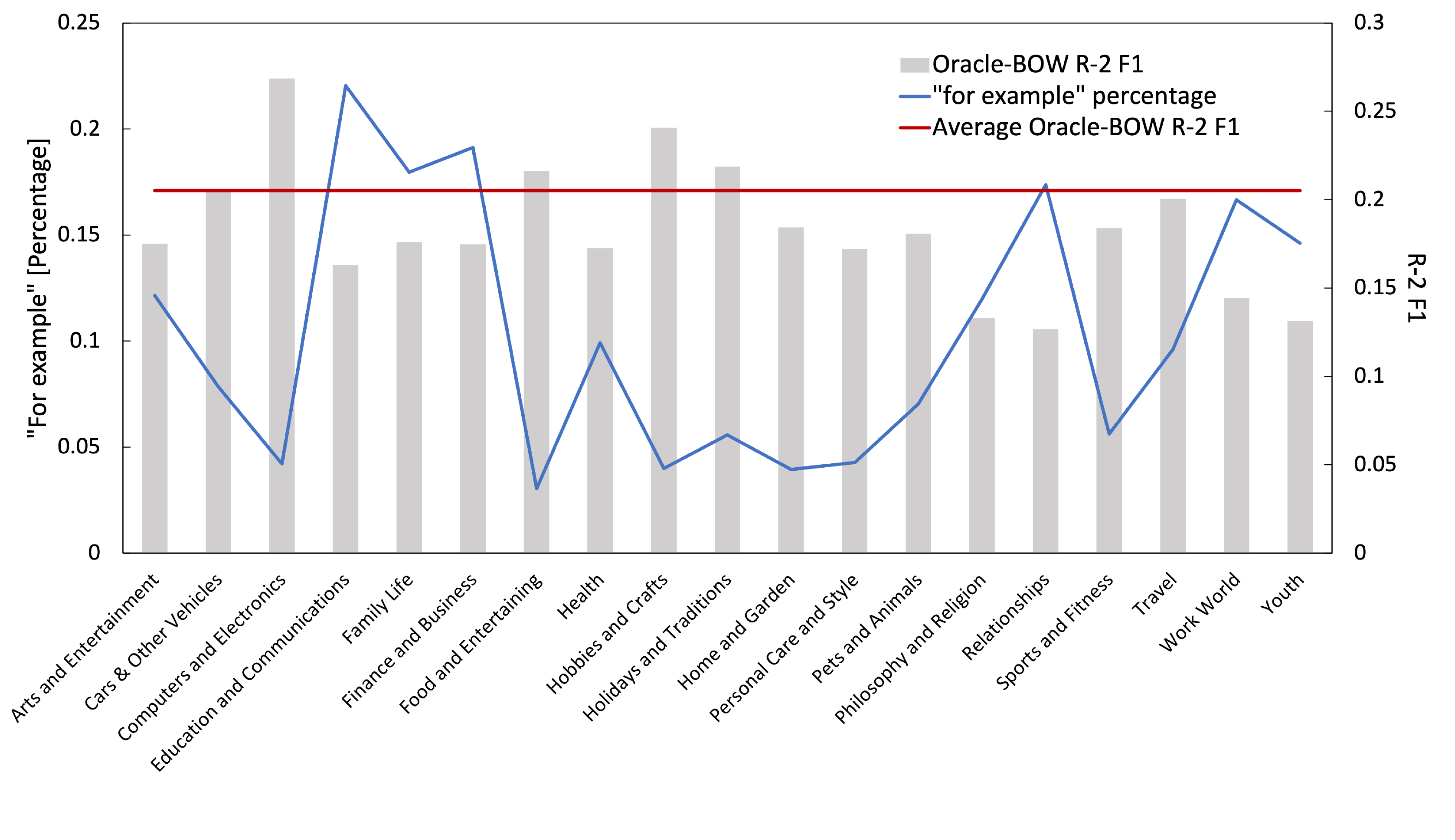}
 \end{center}
  \caption{Examples percentage and ROUGE-2 F-1 of \textsc{Oracle-BOW} per category (\textsc{HowSumm-Step}).}~\label{fig:category-rouge}
\end{figure}

Another observation in section~\ref{sec:summaries} was that \textsc{HowSumm} target summaries are rich with examples. In order to study the potential effect of this phenomenon on extractive models performance we group \textsc{HowSumm-Step} instances by their article's category.  In figure~\ref{fig:category-rouge} we show average ROUGE-2 F-1 score of \textsc{Oracle-BOW} , representing the best extractive output, on these categories. Clearly, the summarization ``potential'' varies between different categories. For example, ``Computers and Electronics'' has an average score of 26.85, which is 50\% higher than the average score on \textsc{HowSumm-Step}. 
On the other hand, ``Relationships'' or ``Youth'' categories did worse than this total average score.  This variation is attributed to authors' original sentences in the target summaries, as these match source sentences poorly. 
Since, according to wikiHow guidelines, examples should not be copied from article's references, we expect categories with low examples percentage to perform better, and vice-versa. 
Indeed, the Pearson correlation between category scores and examples percentage (also depicted in figure~\ref{fig:category-rouge}) is -0.64 showing a moderate negative correlation.


\subsection{Sources Structure Effect}

 The first step in all extractive models is segmenting the sources text into sentences (using Spacy package). Similarly, for abstractive models we split sources' text into paragraphs (using line breaks). 
 These preliminary segmentations of \textsc{HowSumm} sources ignore the fact that they contain many headers of different levels, lists, and Q\&As (see section~\ref{sec:sources}). 
 
 Indeed, the outputs of the extractive models, contain segmentation errors. 
 For example, both header and text are included in ``Raw eggs: There is a risk of food poisoning from raw or partly cooked eggs''. Another example is mashing up all list items in ``Examples of compelling ties include: A residence abroad which you do not intend to abandon Your family relationships Your long term plans''. 
Such errors are also reflected in the Grammaticality score in table~\ref{tab:manual-eval}. 

 Using tools like in ~\cite{heps2015}  
along with strategies to deal with lists, may improve segmentation and headers handling. Thus, allowing to provide higher-quality input to the different models and hopefully better their outputs.

\section{Conclusions}
We introduced a large-scale qMDS summarization dataset based on wikiHow articles  and stressed its uniqueness with regard to other datasets.
In addition, we present ROUGE evaluation results of several extractive and abstractive models on this dataset. The large gap between results of SOTA models and oracles suggests there is room for models improvement using various features of the dataset. Human evaluation hints on the need to improve coherence in the automatic outputs. We believe  that this dataset will promote the development of new qMDS algorithms.

\appendix
\section{Output Examples}\label{examples}
We show in figure ~\ref{figure:summary} some examples of different summarizers outputs, generated for some instances of \textsc{HowSumm-step}.

\begingroup
\renewcommand{\arraystretch}{0.61} 
\begin{figure}[h]
\centering
\resizebox{0.43\textwidth}{!}{%
\begin{tabular}{p{0.7cm}p{6.7cm}}
\multicolumn{2}{c}{{\textit{How to Choose a Medical Alert System for Seniors}}} \\
\hline
\hline
\tiny{\textbf{Method}} & \tiny{\textbf{``Deciding What Level of Monitoring Your Senior Needs''}} \\
\tiny{\textbf{Step}} & \tiny{\textbf{``Consider automatic detection features''}} \\
\hline
\tiny{\textit{Target summary}}& \tiny{Many systems offer a simple push button that allows seniors to call for help whenever they need it. There are also newer products on the market, however, that offer a higher level of protection.  Many companies now offer units with automatic fall detection capabilities, which means your senior will get emergency help even if she is unconscious or otherwise unable to press the emergency button.  Some units are also capable of automatically calling for emergency help if they detect fire or carbon monoxide.}\\
\hline
\hline
\multicolumn{2}{l}{\tiny{\textbf{Summarizers output}}} \\
\hline
\tiny{\textit{HierSumm-BM25}} & \tiny{Look for automatic features like push-button and automatic fall, emergency call center assistance, and medical monitoring.}\\
\hline
\tiny{\textit{HierSumm-Oracle}} & \tiny{Medical alert systems allow senior citizens to live on their own even, since they always have access to help whenever they need it.  For seniors who worry that they may need assistance if they fall down or meet an unexpected emergency, it provides an easy way to get access to help.}\\
\hline
\tiny{\textit{CES}} & \tiny{Some new technologies are addressing the issue by detecting falls automatically.Compared to Monitored Systems If budget allows, it is worthwhile to consider monitored medical alert systems . The system should have the features the senior needs. For an extra monthly fee, customers can get an add-on fall-detection device that will automatically call for help when needed, as long as you have a landline. I would recommend choosing a system that has the ability to upgrade features in the future and that does not have a high cancellation policy.}\\
\hline
\tiny{\textit{LexRank}} & \tiny{Some new technologies are addressing the issue by detecting falls automatically. To appeal to younger seniors and people with active lifestyles, several companies now offer more advanced systems featuring varying combinations of push-button and automatic fall detection, emergency call center assistance, and medical monitoring.A similar product is a pendant-style help button that can automatically place a call for help if it detects a fall. The system should have the features the senior needs.}\\
\hline
\\
\\
\multicolumn{2}{c}{{\textit{How to Adopt As a Single Woman}}} \\
\hline
\hline
\tiny{\textbf{Method}} & \tiny{\textbf{``Submitting Your Application''}} \\
\tiny{\textbf{Step}} & \tiny{\textbf{``Meet with your caseworker.''}} \\
\hline
\tiny{\textit{Target summary}}& \tiny{Once your application is accepted by the agency, you'll be assigned a caseworker who will be your point person throughout the adoption process. They will describe to you the next steps of the process. You need to maintain a good relationship with your caseworker. They are on your side, and want to help you adopt a child. They can answer any questions you have, as well as give you tips and advice.}\\
\hline
\hline
\multicolumn{2}{l}{\tiny{\textbf{Summarizers output}}} \\
\hline
\tiny{\textit{HierSumm-BM25}} & \tiny{Once you have completed your application, you will need to meet with a caseworker who will help you through the adoption process.  In some caseworkers, family workers who work with the children in care will work with the children in care.  you will also need to take time during the application process to get to know them as they will be your guide throughout your adoption journey.}\\
\hline
\tiny{\textit{HierSumm-Oracle}} & \tiny{Once you have completed your application, you will need to meet with your caseworker who will help you through the rest of the adoption process.  In child welfare, generally there are two types of caseworkers  :  family workers who work with families such as yours and child workers who work with children.  It is important to develop and maintain a good relationship with your caseworker, so take time during the application process to get to know them as they will be.}\\
\hline
\tiny{\textit{CES}} & \tiny{Complete Your Application to Adopt This is where the official paperwork begins and where you will meet the caseworker who will help you through the rest of the adoption process.It helps your caseworker understand your family better and assists them with writing your home study.3. Single parent adoptions are also one of the groups that adopts the most special needs children who need families. In this section you will find resources on single-parent adoption from the Child Welfare Information Gateway.}\\
\hline
\tiny{\textit{LexRank}} & \tiny{Complete Your Application to Adopt This is where the official paperwork begins and where you will meet the caseworker who will help you through the rest of the adoption process.It helps your caseworker understand your family better and assists them with writing your home study. It's important to develop and maintain a good relationship with your caseworker, so take time during the application process to get to know them as they will be your guide throughout your adoption journey.}\\
\hline

\end{tabular}
}
\caption{Ground-truth with corresponding automated summaries.}
\label{figure:summary}
\end{figure}
\endgroup


\begin{thebibliography}{24}
\expandafter\ifx\csname natexlab\endcsname\relax\def\natexlab#1{#1}\fi

\bibitem[{Baumel et~al.(2016)Baumel, Cohen, and Elhadad}]{baumel-2016-td-qfs}
Tal Baumel, Raphael Cohen, and Michael Elhadad. 2016.
\newblock Topic concentration in query focused summarization datasets.
\newblock In \emph{Proceedings of the 30th Conference of the Association for
  the Advancement of Artificial Intelligence}, pages 2573--2579.

\bibitem[{Erkan and Radev(2004)}]{lexrank}
G\"{u}nes Erkan and Dragomir~R. Radev. 2004.
\newblock Lexrank: Graph-based lexical centrality as salience in text
  summarization.
\newblock \emph{J. Artif. Int. Res.}, 22(1):457–479.

\bibitem[{Fabbri et~al.(2019)Fabbri, Li, She, Li, and
  Radev}]{fabbri-2019-multiNews}
Alexander Fabbri, Irene Li, Tianwei She, Suyi Li, and Dragomir Radev. 2019.
\newblock \href {https://doi.org/10.18653/v1/P19-1102} {Multi-news: A
  large-scale multi-document summarization dataset and abstractive hierarchical
  model}.
\newblock In \emph{Proceedings of the 57th Annual Meeting of the Association
  for Computational Linguistics}, pages 1074--1084, Florence, Italy.
  Association for Computational Linguistics.

\bibitem[{Feigenblat et~al.(2017)Feigenblat, Roitman, Boni, and
  Konopnicki}]{CES}
Guy Feigenblat, Haggai Roitman, Odellia Boni, and David Konopnicki. 2017.
\newblock \href {https://doi.org/10.1145/3077136.3080690} {Unsupervised
  query-focused multi-document summarization using the cross entropy method}.
\newblock In \emph{Proceedings of the 40th International ACM SIGIR Conference
  on Research and Development in Information Retrieval}, SIGIR '17, page
  961–964, New York, NY, USA. Association for Computing Machinery.

\bibitem[{Gholipour~Ghalandari et~al.(2020)Gholipour~Ghalandari, Hokamp, Pham,
  Glover, and Ifrim}]{wcep-2020}
Demian Gholipour~Ghalandari, Chris Hokamp, Nghia~The Pham, John Glover, and
  Georgiana Ifrim. 2020.
\newblock \href {https://doi.org/10.18653/v1/2020.acl-main.120} {A large-scale
  multi-document summarization dataset from the {W}ikipedia current events
  portal}.
\newblock In \emph{Proceedings of the 58th Annual Meeting of the Association
  for Computational Linguistics}, pages 1302--1308, Online. Association for
  Computational Linguistics.

\bibitem[{Grusky et~al.(2018)Grusky, Naaman, and
  Artzi}]{grusky-etal-2018-newsroom}
Max Grusky, Mor Naaman, and Yoav Artzi. 2018.
\newblock \href {https://doi.org/10.18653/v1/N18-1065} {{N}ewsroom: A dataset
  of 1.3 million summaries with diverse extractive strategies}.
\newblock In \emph{Proceedings of the 2018 Conference of the North {A}merican
  Chapter of the Association for Computational Linguistics: Human Language
  Technologies, Volume 1 (Long Papers)}, pages 708--719, New Orleans,
  Louisiana. Association for Computational Linguistics.

\bibitem[{Koupaee and Wang(2018)}]{DBLP:journals/corr/abs-1810-09305}
Mahnaz Koupaee and William~Yang Wang. 2018.
\newblock \href {http://arxiv.org/abs/1810.09305} {Wikihow: {A} large scale
  text summarization dataset}.
\newblock \emph{CoRR}, abs/1810.09305.

\bibitem[{Kulkarni et~al.(2020)Kulkarni, Chammas, Zhu, Sha, and
  Ie}]{kulkarni2020aquamuse}
Sayali Kulkarni, Sheide Chammas, Wan Zhu, Fei Sha, and Eugene Ie. 2020.
\newblock \href {http://arxiv.org/abs/2010.12694} {Aquamuse: Automatically
  generating datasets for query-based multi-document summarization}.

\bibitem[{Kwiatkowski et~al.(2019)Kwiatkowski, Palomaki, Redfield, Collins,
  Parikh, Alberti, Epstein, Polosukhin, Kelcey, Devlin, Lee, Toutanova, Jones,
  Chang, Dai, Uszkoreit, Le, and Petrov}]{47761}
Tom Kwiatkowski, Jennimaria Palomaki, Olivia Redfield, Michael Collins, Ankur
  Parikh, Chris Alberti, Danielle Epstein, Illia Polosukhin, Matthew Kelcey,
  Jacob Devlin, Kenton Lee, Kristina~N. Toutanova, Llion Jones, Ming-Wei Chang,
  Andrew Dai, Jakob Uszkoreit, Quoc Le, and Slav Petrov. 2019.
\newblock Natural questions: a benchmark for question answering research.
\newblock \emph{Transactions of the Association of Computational Linguistics}.

\bibitem[{Lev et~al.(2020)Lev, Shmueli{-}Scheuer, Jerbi, and
  Konopnicki}]{orgFAQ}
Guy Lev, Michal Shmueli{-}Scheuer, Achiya Jerbi, and David Konopnicki. 2020.
\newblock \href {https://arxiv.org/abs/2009.01460} {orgfaq: {A} new dataset and
  analysis on organizational faqs and user questions}.
\newblock \emph{CoRR}, abs/2009.01460.

\bibitem[{Lin(2004)}]{lin-2004-rouge}
Chin-Yew Lin. 2004.
\newblock \href {https://www.aclweb.org/anthology/W04-1013} {{ROUGE}: A package
  for automatic evaluation of summaries}.
\newblock In \emph{Text Summarization Branches Out}, pages 74--81, Barcelona,
  Spain. Association for Computational Linguistics.

\bibitem[{Liu et~al.(2018)Liu, Saleh, Pot, Goodrich, Sepassi, Kaiser, and
  Shazeer}]{Liu-2018-wikipedia}
Peter~J. Liu, Mohammad Saleh, Etienne Pot, Ben Goodrich, Ryan Sepassi, Lukasz
  Kaiser, and Noam Shazeer. 2018.
\newblock \href {https://openreview.net/forum?id=Hyg0vbWC-} {Generating
  wikipedia by summarizing long sequences}.
\newblock In \emph{6th International Conference on Learning Representations,
  {ICLR} 2018, Vancouver, BC, Canada, April 30 - May 3, 2018, Conference Track
  Proceedings}. OpenReview.net.

\bibitem[{Liu and Lapata(2019)}]{liu-lapata-2019-hierarchical}
Yang Liu and Mirella Lapata. 2019.
\newblock \href {https://doi.org/10.18653/v1/P19-1500} {Hierarchical
  transformers for multi-document summarization}.
\newblock In \emph{Proceedings of the 57th Annual Meeting of the Association
  for Computational Linguistics}, pages 5070--5081, Florence, Italy.
  Association for Computational Linguistics.

\bibitem[{Lu et~al.(2020)Lu, Dong, and Charlin}]{Lu2020MultiXScienceAL}
Yao Lu, Yue Dong, and Laurent Charlin. 2020.
\newblock Multi-xscience: A large-scale dataset for extreme multi-document
  summarization of scientific articles.
\newblock \emph{ArXiv}, abs/2010.14235.

\bibitem[{Manabe and Tajima(2015)}]{heps2015}
Tomohiro Manabe and Keishi Tajima. 2015.
\newblock Extracting logical hierarchical structure of html documents based on
  headings.
\newblock 8(12):1606–1617.

\bibitem[{Nallapati et~al.(2016)Nallapati, Zhou, dos Santos,
  GuÌ‡l{\c{c}}ehre, and Xiang}]{nallapati-etal-2016-abstractive}
Ramesh Nallapati, Bowen Zhou, Cicero dos Santos, {\c{C}}a{\u{g}}lar
  GuÌ‡l{\c{c}}ehre, and Bing Xiang. 2016.
\newblock \href {https://doi.org/10.18653/v1/K16-1028} {Abstractive text
  summarization using sequence-to-sequence {RNN}s and beyond}.
\newblock In \emph{Proceedings of The 20th {SIGNLL} Conference on Computational
  Natural Language Learning}, pages 280--290, Berlin, Germany. Association for
  Computational Linguistics.

\bibitem[{Nenkova and McKeown(2011)}]{nenkova2011}
Ani Nenkova and Kathleen McKeown. 2011.
\newblock \href {https://doi.org/10.1561/1500000015} {\emph{Automatic
  Summarization}}, volume~5.

\bibitem[{Otterbacher et~al.(2005)Otterbacher, Erkan, and
  Radev}]{otterbacher-etal-2005-using}
Jahna Otterbacher, G{\"u}ne{\c{s}} Erkan, and Dragomir Radev. 2005.
\newblock \href {https://www.aclweb.org/anthology/H05-1115} {Using random walks
  for question-focused sentence retrieval}.
\newblock In \emph{Proceedings of Human Language Technology Conference and
  Conference on Empirical Methods in Natural Language Processing}, pages
  915--922, Vancouver, British Columbia, Canada. Association for Computational
  Linguistics.

\bibitem[{Over et~al.(2007)Over, Dang, and Harman}]{10.1016/j.ipm.2007.01.019}
Paul Over, Hoa Dang, and Donna Harman. 2007.
\newblock \href {https://doi.org/10.1016/j.ipm.2007.01.019} {Duc in context}.
\newblock volume~43, page 1506–1520, USA. Pergamon Press, Inc.

\bibitem[{Raffel et~al.(2020)Raffel, Shazeer, Roberts, Lee, Narang, Matena,
  Zhou, Li, and Liu}]{raffel2020exploring}
Colin Raffel, Noam Shazeer, Adam Roberts, Katherine Lee, Sharan Narang, Michael
  Matena, Yanqi Zhou, Wei Li, and Peter~J. Liu. 2020.
\newblock \href {http://arxiv.org/abs/1910.10683} {Exploring the limits of
  transfer learning with a unified text-to-text transformer}.

\bibitem[{Robertson and Zaragoza(2009)}]{10.1561/1500000019}
Stephen Robertson and Hugo Zaragoza. 2009.
\newblock \href {https://doi.org/10.1561/1500000019} {The probabilistic
  relevance framework: Bm25 and beyond}.
\newblock \emph{Found. Trends Inf. Retr.}, 3(4):333–389.

\bibitem[{Rush et~al.(2015)Rush, Chopra, and Weston}]{rush-etal-2015-neural}
Alexander~M. Rush, Sumit Chopra, and Jason Weston. 2015.
\newblock \href {https://doi.org/10.18653/v1/D15-1044} {A neural attention
  model for abstractive sentence summarization}.
\newblock In \emph{Proceedings of the 2015 Conference on Empirical Methods in
  Natural Language Processing}, pages 379--389, Lisbon, Portugal. Association
  for Computational Linguistics.

\bibitem[{Vaswani et~al.(2017)Vaswani, Shazeer, Parmar, Uszkoreit, Jones,
  Gomez, Kaiser, and Polosukhin}]{vaswani2017attention}
Ashish Vaswani, Noam Shazeer, Niki Parmar, Jakob Uszkoreit, Llion Jones,
  Aidan~N Gomez, Lukasz Kaiser, and Illia Polosukhin. 2017.
\newblock Attention is all you need.
\newblock \emph{arXiv preprint arXiv:1706.03762}.

\bibitem[{Zopf(2018)}]{zopf-2018-auto-hmds}
Markus Zopf. 2018.
\newblock \href {https://www.aclweb.org/anthology/L18-1510} {Auto-h{MDS}:
  Automatic construction of a large heterogeneous multilingual multi-document
  summarization corpus}.
\newblock In \emph{Proceedings of the Eleventh International Conference on
  Language Resources and Evaluation ({LREC} 2018)}, Miyazaki, Japan. European
  Language Resources Association (ELRA).

\end{thebibliography}
\end{document}